# Deep Morphological Neural Networks

Yucong Shen, Xin Zhong, and Frank Y. Shih

*Abstract*—Mathematical morphology is a theory and technique to collect features like geometric and topological structures in digital images. Given a target image, determining suitable morphological operations and structuring elements is a cumbersome and time-consuming task. In this paper, a morphological neural network is proposed to address this problem. Serving as a nonlinear feature extracting layer in deep learning frameworks, the efficiency of the proposed morphological layer is confirmed analytically and empirically. With a known target, a single-filter morphological layer learns the structuring element correctly, and an adaptive layer can automatically select appropriate morphological operations. For practical applications, the proposed morphological neural networks are tested on several classification datasets related to shape or geometric image features, and the experimental results have confirmed the high computational efficiency and high accuracy.

## I. Introduction

MATHEMATICAL morphology, based on set theory and topology, is a feature extracting and analyzing technique, where dilation and erosion, i.e., enlarging and shrinking the objects respectively, are the two elementary operations. Applying structuring elements (SE) as sliding windows to identify the feature on digital images, mathematical morphology is typically applied in many core applications in image analysis to extract features like shapes, regions, edges, skeleton, and convex hull [13]. Hence, it facilitates a wide range of applications like defect extraction [4], edge detection [20], and image segmentation [14].

In computer vision, deep learning has become an efficient tool since the development of computer hardware brings an increased computational capability. Many deep learning structures have been proposed based on convolutional neural networks (CNN) [6]. For example, LeNet [8] was proposed for document recognition. Nowadays, deepening the CNN structures enables many computer vision applications, especially for image recognition [18].

In image morphology, given a desired target, it is time-consuming and cumbersome to determine appropriate morphological operations and SE. From the perspective of neural networks and deep learning, researchers have proposed the concept of morphological layers for this issue. Different from the convolutional layers that compute the linear weighted summation in each kernel applied on an image, the dilation layers and erosion layers in morphological neural network (MNN) approximate local maxima and minima and therefore providing nonlinear feature extractors. Similar to the CNN that trains the weights in convolution kernels, MNN intends to learn the weights in SE. Besides, a morphological layer also needs to decide the selection between the operations of dilation and erosion. Since the maximization and minimization operations in morphology are not differentiable, incorporating them into neural networks needs smooth and approximate functions. Ritter *et al.* [11] presented an attempt of MNN formulated by image algebra [12]. Masci *et al.* [10] approximated the dilation and erosion in deep learning framework using counter-harmonic mean. However, only the pseudo dilation and erosion can be achieved since the formulation requires infinite integers. Most recently, Shih *et al.* [17] proposed a deep learning framework to tackle this issue and introduced smooth local maxima and minima layers as the feature extractors. Their MNN roughly approximates dilation and erosion, and learns the binary SE, but cannot learn the non-flat SE and the morphological operations.

In this paper, we propose morphological layers in deep neural networks that can learn both the SE and the morphological operations. Given a target, a morphological layer learns the SE correctly, and an adaptive morphological layer can automatically select appropriate operations by applying a smooth sign function of an extra trainable weight. Considering the strength of morphology in analyzing shape features on images, a residual MNN pipeline and its applications are presented to validate the practicality of the proposed layer. The MNN residual structure is compared against CNN of the same structure on several datasets related to shape features. Experimental results have validated the superior of the proposed MNN in these tasks.

The remainder of this paper is organized as follows. Section II introduces the morphology layers and the residual morphological neural network for shape classifications. Section III presents an adaptive morphological layer for determining the proper morphological operations from the original images and

Y. Shen is with the Computer Vision Laboratory, Department of Computer Science, New Jersey Institute of Technology, Newark, NJ 07102 USA (email: ys496@njit.edu).

X. Zhong is with the Robotics, Networking, and Artificial Intelligence (R.N.A) Laboratory, Department of Computer Science, University of Nebraska at Omaha, Omaha, NE 68106 USA (email: xzhong@unomaha.edu).

F.Y. Shih is with the Computer Vision Laboratory, Department of Computer Science, New Jersey Institute of Technology, Newark, NJ 07102 USA (email: shih@njit.edu).

the desired result images. Section IV shows the experimental results. Finally, conclusions are drawn in Section V.

## II. Deep Morphological Neural Network

### A. The Proposed Morphological Layer

Morphological dilation and erosion are approximated using counter-harmonic mean in [10]. For a grayscale image $f(x)$ and a kernel $\omega(x)$, the core is to define a PConv layer as:

$$PConv(f;\omega,P)(x) = \frac{(f^{P+1}*\omega)(x)}{(f^{P}*\omega)(x)} = (f *_P \omega)(x) \quad (1)$$

where "*" denotes the convolution, and $P$ is a scalar controlling the choice of operation ($P < 0$ is pseudo-erosion, $P > 0$ is pseudo-dilation, and $P = 0$ is standard convolution). Since performing a real dilation or erosion is to require $P$ to be an unachievable infinite number, this formulation can be highly inaccurate in implementation.

Shih *et al.* [17] represented the dilation and erosion using the soft maximum and soft minimum functions. In a dilation layer, the $j$-th pixel in the $s$-th feature map $z \in \mathbb{R}^n$ is

$$z_j^s = \ln(\sum_{i=1}^{n} e^{\omega_i x_i}) \quad (2)$$

where $n$ is the total number of weights in a SE, $x_i$ is the $i$-th element of the masked window in the input image, and $\omega_i$ is the $i$-th element of the current weight. The $z$ is similar in an erosion layer as

$$z_j^s = -\ln(\sum_{i=1}^{n} e^{-\omega_i x_i}) \quad (3)$$

Although Shih *et al.* [17] approximates dilation and erosion theoretically, it failed to learn the SE accurately. Training on the samples of input images and desired morphological images, a single-layer and single-filter MNN always miss some elements on the SE. Fig. 1 presents the architecture of the single-layer MNN and Fig. 2 shows some SE learned, from which we can observe the errors. These missing points are caused by rounding errors. Eqs. (2) and (3) do not round the floating points when computing the maxima and minima in a sliding window, while the neural network tries to minimize the difference between the predicted and the target images. As a result, the learned SE compensates these floating points and hence contains the rounding errors. The notations and terms used in this paper are listed in Table I.

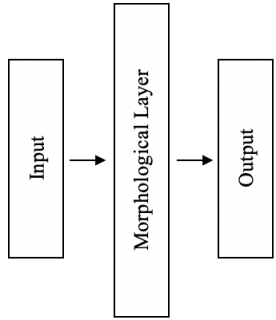

Fig. 1. Architecture of the single layer MNN.

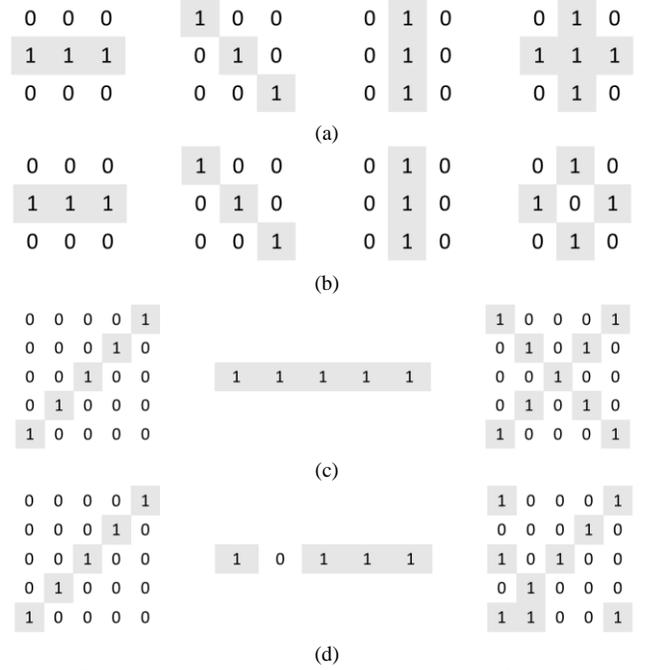

Fig. 2. (a) The horizontal, diagonal, vertical, and diamond $3 \times 3$ structuring elements applied to input images when creating target images, (b) the corresponding structuring elements learned by the single dilation layer MNN, (c) the original 45°, crossing $5 \times 5$ structuring elements and horizontal line $1 \times 5$ structuring elements applied on the inputs images when creating target images, (d) the corresponding structuring elements learned by the single dilation layer MNN.

TABLE I
NOTATIONS AND TERMS USED IN THIS PAPER

| | |
|---|---|
| $\omega$ | The weight of morphological layer. |
| $x$ | The input image of morphological layer. |
| $b$ | The bias matrix. |
| $n = a \times b$ | The total number of weights in a SE, $a$ is the width and $b$ is the height of SE. |
| $W_i$ | The $i$-th element of the SE $W$. |
| $X_i$ | The $i$-th element of the current masked window in the original image $X$. |
| $\hat{y}$ | The output of the network. |
| $y$ | The target of the network. |

Definition 1: The *differentiable binary dilation* of the $j$-th pixel in an output image $Y \in \mathbb{R}^n$ is defined as

$$Y_j = \ln(\sum_{i=1}^{n} e^{W_i X_i}) \quad (4)$$

where $W$ is a binary SE slid on the input image $X$, and the default stride is 1. We denote it as $W \oplus X$, where $W \in \mathbb{R}^n$ and $X \in \mathbb{R}^n$.

Definition 2: The *differentiable binary erosion* of the $j$-th pixel in an output image $Y \in \mathbb{R}^n$ is defined as

$$Y_j = -ln(\sum_{i=1}^{n} e^{-W_i X_i}) \quad (5)$$

where $W$ is the binary SE slid on the input image $X$, and the default stride is 1. We denote it as $W \ominus X$, where $W \in \mathbb{R}^n$ and $X \in \mathbb{R}^n$.

When the binary dilation is learned,

$$ln(\sum_{i=1}^{n} e^{W_i X_i}) \geq max\ (W_i X_1, W_2 X_2, \ldots, W_n X_n) \quad (6)$$

which indicates that

$$ln(\sum_{i=1}^{n} e^{W_i X_i}) \geq X_i. \quad (7)$$

Therefore, we have

$$\sum_{i=1}^{n} e^{W_i X_i} \geq e^{X_i}. \quad (8)$$

Clearly, Eq. (8) is invalid. To tackle this issue, a slack variable $\zeta$ is added to obtain

$$\sum_{i=1}^{n} e^{W_i X_i} \zeta \geq e^{X_i}. \quad (9)$$

Note that Eq. (9) is valid when $\zeta \geq \frac{e^{X_i}}{\sum_{i=1}^{n} e^{W_i X_i}}$. Similarly, a slack variable can be applied to validate the differentiable binary erosion.

Inspired by the CNN and Eq. (9), we apply bias variables to correct the rounding errors caused by the soft maximum and soft minimum functions. Different from the traditional way of applying one bias number in each filter, our bias is a matrix of the same size as the input image to correct the error of each point. In a binary dilation layer, the $s$-th feature map $z$ of a binary dilation layer will be

$$z^s = \omega \oplus x + b \quad (10)$$

where $\omega \in \mathbb{R}^n$, $x \in \mathbb{R}^n$, and $b \in \mathbb{R}^n$.

The $s$-th feature map $z$ in a binary erosion layer is

$$z^s = \omega \ominus x + b. \quad (11)$$

After adding $b$ in Eq. (10), we can obtain

$$(\sum_{i=1}^{n} e^{W_i X_i}) \cdot e^b \geq e^{X_i}. \quad (12)$$

Note that Eq. (12) is valid when $b \geq ln \frac{e^{X_i}}{\sum_{i=1}^{n} e^{W_i X_i}}$. Therefore, dilation layer is correct if $b \geq ln \frac{e^{X_i}}{\sum_{i=1}^{n} e^{W_i X_i}}$ after training. We can derive the correctness condition for an erosion layer similarly.

Definition 3: The *differentiable grayscale dilation* of the $j$-th pixel in an output image $Y \in \mathbb{R}^n$ is defined as

$$Y_j = ln(\sum_{i=1}^{n} e^{W_i + X_i}) \quad (13)$$

where $W$ is the non-flat SE slid on the input image $X$, and the default stride is 1. We denote it as $W \oplus_g X$, where $W \in \mathbb{R}^n$, and $X \in \mathbb{R}^n$.

Definition 4: The *differentiable grayscale erosion* of the $j$-th pixel in an output image $Y \in \mathbb{R}^n$ is defined as

$$Y_j = -ln(\sum_{i=1}^{n} e^{-(W_i - X_i)}) \quad (14)$$

where $W$ is the non-flat SE slid on the input image $X$, and the default stride is 1. We denote it as $W \ominus_g X$, where $W \in \mathbb{R}^n$, and $X \in \mathbb{R}^n$.

When learning a dilation with a non-flat SE like Eq. (6), there should have

$$ln(\sum_{i=1}^{n} e^{W_i + X_i}) \geq max\ (W_i + X_1, \ldots, W_n + X_n). \quad (15)$$

We can obtain

$$\sum_{i=1}^{n} e^{W_i + X_i} \geq e^{W_i + X_i}. \quad (16)$$

Clearly Eq. (16) is valid. Therefore, we can prove the correctness of differentiable grayscale dilation. Like in the binary dilation layer, we apply a bias vector to correct the rounding errors caused by the soft maximum and soft minimum functions. The $s$-th feature map $z^s$ of a grayscale dilation layer is

$$z^s = \omega \oplus_g x + b \quad (17)$$

where $\omega \in \mathbb{R}^n$, $x \in \mathbb{R}^n$, and $b \in \mathbb{R}^n$.

In grayscale erosion layer, the $s$-th feature map $z^s$ is

$$z^s = \omega \ominus_g x + b \quad (18)$$

where $\omega \in \mathbb{R}^n$, $x \in \mathbb{R}^n$, and $b \in \mathbb{R}^n$.

The gradient of the proposed morphological layer is computed by the back-propagation [8] following the chain rule. Denoted the objective function as $J(\omega, b; y, \hat{y})$, the gradient $\delta^{(l)}$ of the $l$-th layer of the network with respect to weight $\omega$ is

$$\delta^{(l)} = \frac{\partial J(\omega, b; y, \hat{y})}{\partial \omega^{(l)}}. \quad (19)$$

Assuming that the learning rate is $\eta$, the weight $\omega$ of the $l$-th layer in $t$-th iteration is updated as

$$\omega_{t+1}^{(l)} = \omega_t^{(l)} - \eta \delta_t^{(l)}. \quad (20)$$

The bias $b$ is updated as

$$b_{t+1}^{(l)} = b_t^{(l)} - \eta \frac{\partial J(\omega, b; y, \hat{y})}{\partial b_t^{(l)}}. \quad (21)$$

## B. Deep MNN with Stacked Morphological Layers

The target images with multiple morphological operations can be learned by stacking the morphological layers to construct a multi-layer MNN. Assuming that the $l$-th layer of multi-layer MNN is a dilation layer, the $s$-th feature map $z_s^{(l)} \in \mathbb{R}^n$ is

$$z_s^{(l)} = \omega \oplus z^{(l-1)} + b \qquad (22)$$

where $\omega \in \mathbb{R}^n$, and $z^{(l-1)} \in \mathbb{R}^n$ is the output of $(l-1)$-th layer.

If the $l$-th layer of multi-layer MNN is an erosion layer, the $s$-th feature map of the output $z \in \mathbb{R}^n$ of current layer will become:

$$z_s^{(l)} = \omega \ominus z^{(l-1)} + b \qquad (23)$$

where $\omega \in \mathbb{R}^n$.

Fig. 3 shows the architecture of the multi-layer deep MNN. The inputs are the original images, and the outputs are the predictions of network after multiple morphological layers. The target images are created by a sequence of morphological operations. At convergence, the deep MNN can learn the SE that minimizes the distance between the input and the target images.

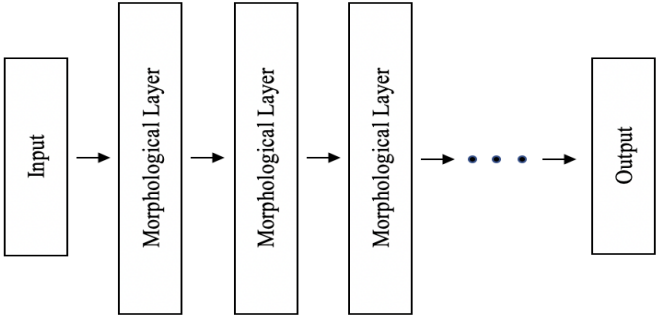

Fig. 3. Architecture of the multi-layer deep morphological neural network.

The gradient of multi-layer DMNN is computed by back-propagation with chain rule. Let the objective function be denoted as $J(\omega, b; y, \hat{y})$. The gradient $\delta^{(l)}$ of the $l$-th layer with respect to weight $\omega$ can be expressed as

$$\delta^{(l)} = \frac{\partial J(\omega,b;y,\hat{y})}{\partial \omega^{(l)}} = \frac{\partial J(\omega,b;y,\hat{y})}{\partial z^{(l)}} \frac{\partial}{\partial \omega} \sigma(z^{(l)}) \qquad (24)$$

where $\sigma(\cdot)$ is the activation function. Assuming that the learning rate is $\eta$, the weight $\omega$ of the $l$-th layer in iteration $t$ is updated by

$$\omega_{t+1}^{(l)} = \omega_t^{(l)} - \eta \delta^{(l)}. \qquad (25)$$

## C. Residual MNN

Mathematical morphology is designed to deal with shapes and structures [5, 19] in applications. In pattern recognition, mathematical morphology is used for preprocessing and feature extraction. In this residual MNN, we aim to apply opening with circular SE on the original image to round the corners of a shape and subtract the rounded-corner image from the original image. The morphological residuals that indicate the corners of a shape can help classification. Fig. 4 shows an example of a morphological residual model.

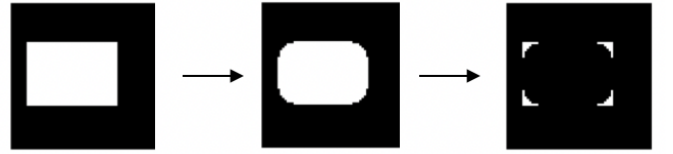

Fig. 4. The morphological residual model. Applying opening on the original image with circle structuring elements, then subtraction of result image from original image can obtain the morphological residuals.

We construct this residual MNN for shape classification as shown in Fig. 5. The input of the neural network is batches of images and an erosion layer followed by a dilation layer means applying an opening on the input images. After the subtraction layer, the neural network finishes the preprocessing progress and delivers the residuals to a classifier. There are two fully-connected layers to take the votes of each pixel in the residuals.

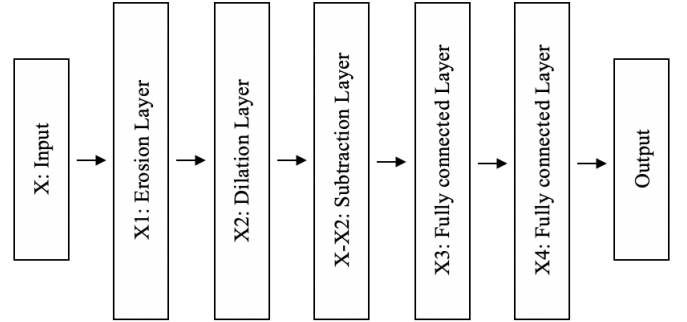

Fig. 5. The architecture of residual MNN.

The configuration of the residual MNN is shown in Table II. The number of channels $m$ should be the same in each layer.

TABLE II
THE CONFIGURATION OF RESIDUAL MORPHOLOGICAL NEURAL NETWORK

| | Input |
|---|---|
| 1 | Erosion $3 \times 3 \times m$ |
| 2 | Dilation $3 \times 3 \times m$ |
| 3 | Subtraction $m$ |
| 4 | FC-1024 |
| 5 | FC-512 |
| 6 | Soft-max |

The residual MNN can be trained by back-propagation. The weights of dilation layer, erosion layer and fully-connected layers are updated by Eqs. (24) and (25). The weights are not updated in the subtraction layer, i.e., the residual MNN just transmits the gradient from the fourth layer to the second layer.

Assuming that the gradient of the fourth layer is $\delta^{(4)}$, the gradient of subtraction layer is $\delta^{(3)} = \delta^{(4)}$.

### III. ADAPTIVE MORPHOLOGICAL LAYER

Determining appropriate operation is a crucial task in the proposed MNN. Among various morphological operations, such as dilation, erosion, opening, and closing, the proposed layer aims to make a decision on the atomic ones including dilation and erosion.

Obviously, the difference between the differentiable binary dilation and the differentiable binary erosion is the sign before the weights. Therefore, we can apply a sign function to choose between maximum and minimum. To make the sign trainable, an extra weight is introduced in the MNN kernels. We call such a morphological layer with this extra weight as an *adaptive morphological layer*. Mathematically, the $j$-th pixel on the output $z \in \mathbb{R}^n$ of the dilation and erosion layers can be represented by

$$z_j = sign(a) \cdot ln(\sum_{i=1}^{n} e^{sign(a) \cdot \omega_i x_i}) + b \quad (26)$$

where $a$ is an extra trainable variable aside from $\omega_i$ and $b$. If $sign(a)$ is $+1$, the operation of current layer is dilation; if $sign(a)$ is $-1$, the operation of current layer is erosion. Since the sign function is not a continuous function and not differentiable, it cannot be used in the neural network. Therefore, we adopt a smooth sign function in the interval $[-1,1]$. Eqs. (27) and (28) show the soft sign function and the hyperbolic tangent function, respectively.

$$f(x) = \frac{x}{1+|x|} \quad (27)$$

$$g(x) = \frac{e^x - e^{-x}}{e^x + e^{-x}} \quad (28)$$

We replace the sign function in Eq. (26) by the hyperbolic tangent function and the soft sign function to maintain the gradient flow. Then the $j$-th pixel on the output $z \in \mathbb{R}^n$ of the adaptive morphological layer is computed in two ways:

$$z_j = \frac{a}{1+|a|} \cdot ln(\sum_{i=1}^{n} e^{\frac{a}{1+|a|} \cdot \omega_i x_i}) + b \quad (29)$$

or

$$z_j = \frac{e^a - e^{-a}}{e^a + e^{-a}} \cdot ln(\sum_{i=1}^{n} e^{\frac{e^a - e^{-a}}{e^a + e^{-a}} \cdot \omega_i x_i}) + b \quad (30)$$

where $a$ is a trainable variable and $a \in \mathbb{R}$.

Fig. 6 compares the soft sign function and the hyperbolic tangent function. The hyperbolic tangent function reaches $-1$ and $+1$ ahead of the soft sign function in that the value of soft sign function is around $-0.5$ when tanh function reaches $-1$. Similarly, the value of the soft sign function lies around $0.5$ when tanh function almost reaches $+1$. Therefore, the gradient of the soft sign is always smaller than the hyperbolic tangent function. In conclusion, the hyperbolic tangent function outperforms the soft sign function theoretically.

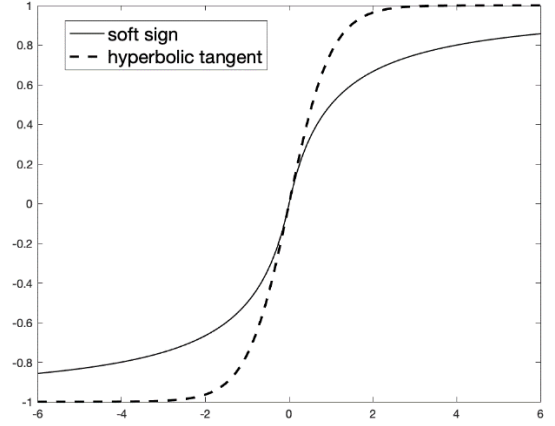

Fig. 6. The soft sign function and hyperbolic tangent function.

A single-layer MNN with the adaptive morphological layer is constructed to test performance. The input is the original images, and the target images are dilated or eroded images. The proposed adaptive morphological layer successfully learns both the target and the choice between dilation and erosion. Fig. 7 shows the flow chart of detecting morphological operations by a single adaptive morphological layer MNN. The MNN minimizes the distance between target and output images. After converges, if the smooth sign function is $+1$, the target images are dilated images; if the smooth sign function is $-1$, the target images are eroded images.

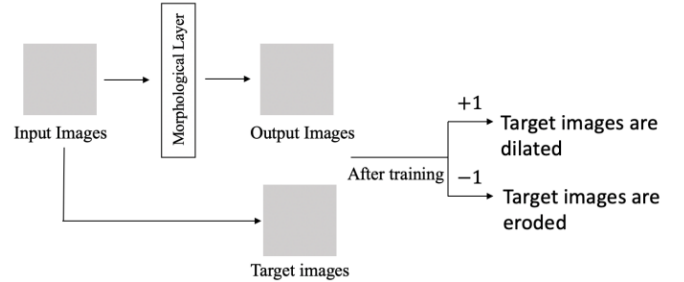

Fig. 7. The flow chart of detecting morphological operations by a single adaptive morphological layer MNN.

In the adaptive layer, the gradients are updated by back-propagation with chain rule. The weight is updated by gradient descent for optimization. Let the objective function of such a neural network be $J(\omega, b, a; y, \hat{y})$. The gradient $\delta^{(l)}$ of the $l$-th layer with respect to weight $a$ is:

$$\delta^{(l)} = \frac{\partial J(\omega,b,a;y,\hat{y})}{\partial a^{(l)}} = \frac{\partial J(\omega,b,a;y,\hat{y})}{\partial z^{(l)}} \frac{\partial z^{(l)}}{a^{(l)}} = \frac{\partial J(\omega,b,a;y,\hat{y})}{\partial z^{(l)}} \varphi'(a) \quad (31)$$

where $\varphi(\cdot)$ is the soft sign or the hyperbolic tangent function.

Assuming that the learning rate is $\eta$, the weight $a$ of the $l$-th layer in $t$-th iteration is updated by

$$a_{t+1}^{(l)} = a_t^{(l)} - \eta \delta^{(l)}. \quad (32)$$

## IV. Experimental Results

The experiments are performed on a 4 NVIDIA Titan X GPU system. We present our experimental results on four datasets including MNIST, a self-created geometric shapes (SCGS) dataset, a German Traffic Sign Recognition Benchmark (GTSRB) dataset [16], and a brain tumor dataset [1], to highlight the strength of the proposed MNN in analyzing shape features.

MNIST dataset is a database consisting of 70,000 examples of handwritten digits 0~9. It has 60,000 training images and 10,000 testing images. They are all $28 \times 28$ grayscale images in 10 classes. The SCGS dataset contains 120,000 grayscale images of size $64 \times 64$ in 5 classes: ellipse, line, rectangle, triangle, and five-edge polygon. The images are created by randomly drawing white objects on a black background, where the size, position, and orientation are randomly initialized. There are 20,000 images in each class for training and 5,000 images used in each class for testing. GTSRB introduces a single-image, multi-class classification problem, and there are 42 classes in total. The images contain one traffic sign each, and each real-world traffic sign only occurs once. We resize all the images into $31 \times 35$, and select 31,367 images for training and 7,842 images for testing. All the images are converted to grayscale. The MRI Brain Tumor Dataset [1] contains 3,064 grayscale T1-weighted contrast-enhanced images from 233 patients with three kinds of brain tumor: meningioma (708 samples), glioma (1426 samples), and pituitary tumor (930 samples). We resize all the images into $64 \times 64$ for classification, and 2,910 images are used for training and 154 images for testing. Fig. 8 shows some examples from the four datasets.

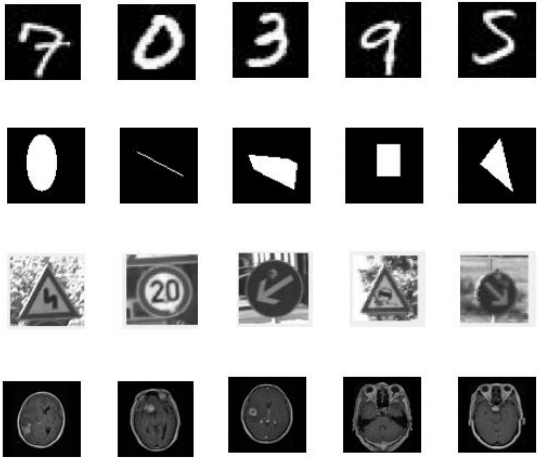

Fig. 8. The examples from the four datasets in the experiments. The first row is the images from MNIST dataset, the second row from SCGS dataset, the third row from GTSRB dataset, and the fourth row from brain tumor dataset.

### A. Learning the SE and Morphological Targets

We randomly select 10,000 images from MNIST dataset, and construct a single-layer MNN as shown in Fig. 1 to learn a single binary SE. Mean squared error (MSE) is adopted to measure the distance between the target and predicted images to minimize the number outliers. The target images are created by applying dilation or erosion on the original input images. A mini-batch SGD [9] with a batch size of 64 and the learning rate $\eta = 7.50$ are selected. When learning binary SE, we adopt three examples, $3 \times 3$ diamond SE, $5 \times 5$ crossing SE, and $1 \times 5$ horizontal line SE. The experiment is repeated 100 times by randomly selecting 10,000 training images each time. The single-layer MNN has 100% accuracy on learning the $3 \times 3$ diamond SE and the $1 \times 5$ horizontal line SE, and 91% accuracy on the $5 \times 5$ crossing SE. Furthermore, if we increase the epochs from 20 to 100, the accuracy on learning the $5 \times 5$ crossing SE is increased to 97%. Fig. 9 shows three examples, where the learned and the original structuring elements are identical.

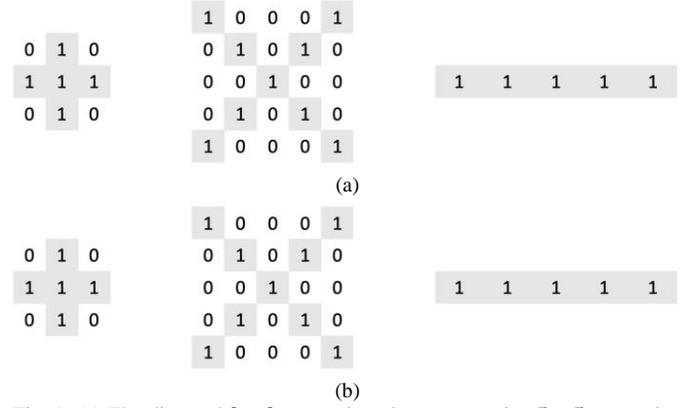

Fig. 9. (a) The diamond $3 \times 3$ structuring element, crossing $5 \times 5$ structuring element, and $1 \times 5$ structuring element, (b) the learned structuring elements by a single dilation layer MNN after improvement.

All the settings in the grayscale morphology experiment are the same as the binary SE case, except that the targets are created by applying non-flat SE, the learning rate is $\eta = 1.0$ for learning dilated images, and $\eta = 0.5$ for learning eroded images.

We adopt the taxicab distance to measure the distance between the learned non-flat SE and the applied non-flat SE after training. Let two matrices be $\boldsymbol{A} = (a_{ij})$ and $\boldsymbol{B} = (b_{ij})$. The taxicab distance between $\boldsymbol{A}$ and $\boldsymbol{B}$ is computed as

$$d_1(\boldsymbol{A}, \boldsymbol{B}) = \sum_{i=1}^{n} \sum_{j=1}^{n} |a_{ij} - b_{ij}| \tag{34}$$

For learning grayscale dilation and erosion, the MNN is converged in 20 epochs. For learning the morphological SE, 100 experiments are done on the same dataset to obtain the distance between the $3 \times 3$ SE and the learned SE. The average distance is 0.0706 in learning dilation and 0.0875 in learning erosion. For learning the morphological targets, the MSE is around $3.43 \times 10^{-5}$ in dilation, and $7.59 \times 10^{-5}$ in erosion.

Fig. 10 show some examples of learning a non-flat SE. It is observed that the original non-flat SE and the learned SE are very close. Fig. 11 shows the target images and the prediction of the network.

| 0.2060 | 0.3234 | 0.6542 |     | 0.8329 | 0.4865 | 0.9737 |
|--------|--------|--------|-----|--------|--------|--------|
| 0.3551 | 0.5692 | 0.3950 |     | 0.0440 | 0.8055 | 0.1752 |
| 0.6405 | 0.5834 | 0.5104 |     | 0.6563 | 0.5816 | 0.0463 |

| 0.2086 | 0.3211 | 0.6521 |     | 0.8361 | 0.4876 | 0.8951 |
|--------|--------|--------|-----|--------|--------|--------|
| 0.3540 | 0.8055 | 0.4135 |     | 0.0559 | 0.5054 | 0.1763 |
| 0.6261 | 0.5747 | 0.5097 |     | 0.6585 | 0.5836 | 0.0573 |

(a)          (b)

Fig. 10. (a) The top box shows the original structuring element and the bottom box shows the learned structuring elements by a single dilation layer MNN. (b) The top box shows the original structuring elements and the bottom box shows the learned structuring elements by a single erosion layer MNN.

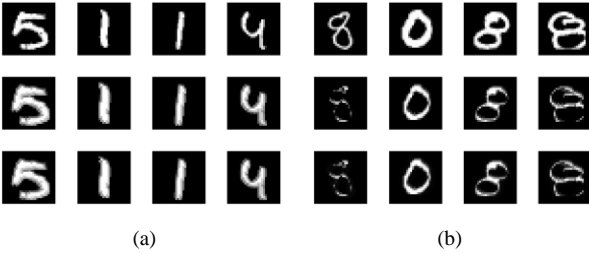

(a)          (b)

Fig. 11. The results of learning grayscale (a) dilation and (b) erosion operations by MNN. The first row shows the original images, the second row shows the target images, and the third row shows the output of the network after training 20 epochs.

In mathematical morphology, opening and closing are also important, where the opening is an erosion followed by a dilation, and conversely, the closing is a dilation followed by an erosion. Therefore, we construct a two-layer MNN to learn the opening or closing. The corresponding targets are the opened or the closed image, and the MNN consists of two layers of dilation after erosion or erosion after dilation. The mini-batch SGD with a batch size of 64 and the learning rate $\eta = 10.0$ are selected. The loss is converged at around 0 within 10 epochs. Fig. 12 shows some examples.

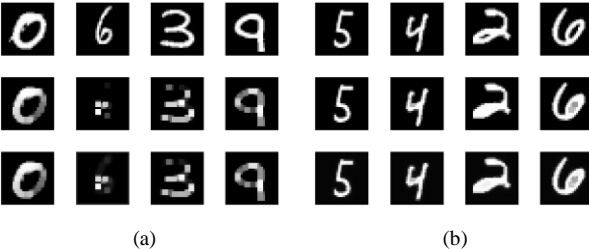

(a)          (b)

Fig. 12. The results of learning (a) opening and (b) closing operations by DMNN. The first row shows the original images, the second row shows the target images, and the third row shows the output of the network after training 20 epochs.

### B. Learning the Morphological Operations

We choose 10,000 images from the MNIST dataset randomly for learning the morphological operations. The MNN consists of a single adaptive morphological layer that minimizes the distance between dilated (or eroded) and the predicted images. At the convergence, the extra weight $a$ in Eqs. (29) and (30) indicates the operation by its sign. The target images are predicted as dilated images if soft sign or hyperbolic tangent function is rounded to $+1$ and as eroded images if $-1$.

The mini-batch SGD is used to optimize the network. The batch size of 64 and the learning rate $\eta = 10.0$ are selected. The distance between the predicted and target images is measured by MSE loss. After 20 epochs, the single adaptive MNN converges, and the MSE loss decreases to around $3 \times 10^{-4}$. If the value of the smooth sign function larger than 0.5, we round it to 1; if the value of the smooth sign function smaller than -0.5, we round it to -1. The values of smooth sign function are in the interval $[-1,1]$. The experiment is repeated 100 times by randomly selected 10,000 images from MNIST each time. Table III shows the detection accuracy of dilation/erosion by two smooth sign functions.

TABLE III
DETECTION ACCURACY OF TWO SMOOTH SIGN FUNCTIONS

|                    | Dilation | Erosion |
|--------------------|----------|---------|
| Soft sign          | 100%     | 100%    |
| Hyperbolic tangent | 100%     | 100%    |

### C. Classification

The mini-batch algorithm with a batch size 64 and learning rate $\eta = 0.0001$ are selected. The residual MNN converges within 100 epochs for all the datasets. The testing accuracy of the residual MNN is 98.93% on MNIST dataset, 98.89% on self-created geometric shape dataset, 95.35% on GTSRB, and 95.43% on MRI brain tumor dataset. We add a dropout layer after the second fully-connected layer to prevent the overfitting when training on the GTSRB, the testing accuracy increases to 96.49%. Table IV shows the configurations of the residual MNN when training on four datasets, where $a$ indicates the number of filters applied in each layer.

TABLE IV
STRUCTURES OF RESIDUAL MNN ON FOUR DATASETS

|                       | MNIST               | SCGS                | GTSRB               | Brain tumor         |
|-----------------------|---------------------|---------------------|---------------------|---------------------|
| Erosion layer         | $3 \times 3 \times a$ | $3 \times 3 \times a$ | $3 \times 3 \times a$ | $3 \times 3 \times a$ |
| Dilation layer        | $3 \times 3 \times a$ | $3 \times 3 \times a$ | $3 \times 3 \times a$ | $3 \times 3 \times a$ |
| Subtraction layer     | $28 \times 28 \times a$ | $64 \times 64 \times a$ | $31 \times 35 \times a$ | $64 \times 64 \times a$ |
| Fully-connected layer | 120                 | 1024                | 1024                | 512                 |
| Fully-connected layer | 84                  | 512                 | 512                 | N/A                 |
| Output                | 10                  | 5                   | 43                  | 3                   |

To quantitatively compare the residual MNN against other methods, we add one more convolutional layer to extract more features and decrease the size of the filters from $5 \times 5$ to $3 \times 3$ in LeNet and name it as Modified LeNet (MLeNet). Table V shows the configuration of MLeNet.

TABLE V
CONFIGURATION OF MLENET

|   | Input                              |
|---|------------------------------------|
| 1 | Convolutional layer $3 \times 3 \times 16$ |
| 2 | Max pooling 2 × 2 |
| 3 | Convolutional layer 3 × 3 × 32 |
| 4 | Max pooling 2 × 2 |
| 5 | Convolutional layer 3 × 3 × 64 |
| 6 | Max pooling 2 × 2 |
| 7 | Fully-connected 2048 × 1 |
| 8 | Fully-connected 1024× 1 |
| 9 | Softmax |

TABLE VI
COMPARISON OF RESIDUAL MNN WITH STATE-OF-ART CONVOLUTIONAL NEURAL NETWORKS

| Classifier | Dataset | Testing accuracy | Number of parameters |
|---|---|---|---|
| MCDNN [3] | MNIST | 99.77% | 2,682,470 |
| Residual MNN | MNIST | 98.93% | 104,181 |
| MLeNet | SCGS | 99.50% | 10,493,795 |
| Residual MNN | SCGS | 98.89% | 4,721,175 |
| MLeNet | GTSRB (Grayscale) | 97.94% | 4,202,339 |
| Residual MNN | GTSRB (Grayscale) | 96.49% | 1,594,903 |
| MLeNet | Brain tumor | 96.10% | 10,493,795 |
| Residual MNN | Brain tumor | 95.43% | 4,721,175 |

Table VI shows the comparisons between the residual MNN against some state-of-the-art CNNs when $a = 1$. Although the residual MNN loses on the testing accuracy as compared to some of state-of-the-art CNNs, it has much less parameters. Especially in the feature extraction layers, the residual MNN has only 20 parameters in total, while the CNNs have at least thousands of parameters. We also show the comparison of the number of parameters in feature extraction layer of residual MNN with state-of-art CNN in Table VII. From Tables VI and VII, we conclude that residual MNN uses much less parameters in feature extraction layers without significantly compromising the model accuracy. The residual MNN has great a tradeoff between the computational efficiency and testing accuracy.

TABLE VII
COMPARISON OF NUMBER OF PARAMETERS IN FEATURE EXTRACTION LAYER OF RESIDUAL MNN WITH STATE-OF-ART CNN

| Model | Number of parameters in feature extraction layers |
|---|---|
| Residual MNN | 20 |
| MLeNet | 2,912 |
| MCDNN | 739,900 |

To further demonstrate the advantages of the proposed morphological layers, we construct a CNN that has the same configuration with the residual MNN and compare their performance on shape related classification. Table VIII shows the configuration of the CNN, which is named as residual CNN, where $b$ denotes the number of filters in each layer.

TABLE VIII
CONFIGURATION OF RESIDUAL CNN

| | Input |
|---|---|
| 1 | Convolutional layer 3 × 3 × $b$ |
| 2 | Convolutional layer 3 × 3 × $b$ |
| 3 | Subtraction layer 3 × 3 × $b$ |
| 4 | Fully-connected 2048 × $b$ |
| 5 | Fully-connected 1024× $b$ |
| 6 | Softmax |

Table IX shows another comparison of the residual CNN against the residual MNN on classifying the four datasets. When working on the brain tumor dataset, we remove the fourth layer shows in Table VIII to keep the configuration of residual CNN as same as the residual MNN in the fifth column of Table IV.

TABLE IX
COMPARISON OF RESIDUAL MNN AND RESIDUAL CNN

| | Residual MNN ($a = 1$) | Residual CNN ($b = 1$) | Residual MNN ($a = 16$) | Residual CNN ($b = 16$) |
|---|---|---|---|---|
| MNIST | 98.93% | 97.14% | 97.78% | 98.18% |
| SCGS | 98.89% | 98.25% | 98.90% | 98.91% |
| GTSRB | 96.49% | 90.60% | 97.48% | 93.39% |
| Brain tumor | 95.43% | 96.10% | 96.75% | 94.15% |

In table IX, when $a = 1$ and $b = 1$, the residual MNN has better testing accuracy on all these datasets than the residual CNN. When a= 16 and $b = 16$, the residual MNN has better testing accuracy on GTSRB dataset. In brain tumor dataset, the residual MNN performs better when $a = 16$, but loses a little bit when $a = 1$. The morphological layers perform better on MRI brain tumor dataset when each layer has multiple filters. Therefore, morphological layer outperforms convolutional layers if both neural networks have same structure in general. Especially on GTSRB dataset that is closely related to shape features, morphological layer significantly improves the testing accuracy, which indicates that proposed MNN has advantages in shape feature extractions.

Therefore, with same number of parameters, the residual MNN loses a little bit on MNIST dataset, but performs the best on GTSRB and brain tumor datasets which are related to shape features. In general, morphological layer works better on extracting shape features than convolutional layer when the residual MNN has same number of parameters as residual CNN.

In summary, MNN has better accuracy in the selected datasets when it has the same number of parameters and the same structures with the CNN. In addition, MNN significantly saves parameters when it has similar accuracy with the CNN. The proposed residual MNN provides a tradeoff between model accuracy and model complexity.

## V. CONCLUSIONS

We propose deep morphological neural networks in this paper that can learn both the operations and corresponding structuring elements in mathematical morphology. The promising performance of the proposed morphological layers, serving as an effective non-linear feature extractor, is confirmed theoretically and experimentally. We also present an architecture of residual MNN for the feature extraction in shape classification tasks to validate the practicality of our morphological neural networks, which shows its superior by providing a good tradeoff between model accuracy and computational complexity.

REFERENCES

[1] J. Cheng, "Brain tumor dataset," *figshare*, 02, Apr. 2017.
[2] R. Coliban, M. Ivanovici and N. Richard, "Improved probabilistic pseudo-morphology for noise reduction in colour images," in *IET Image Process.*, vol. 10, no. 7, pp. 505-514, 7 2016.
[3] D. Ciregan, U. Meier and J. Schmidhuber, "Multi-column deep neural networks for image classification," in *Proc. IEEE Conf. Comput. Vis. Pattern Recognit.*, Providence, Rhode Island, 2012, pp.3642-3649.
[4] K. Chen, Z. Zhang, Y. Chao, M. Dai and J. Shi, "Defects extraction for QFN based on mathematical morphology and modified region growing," in *Proc. IEEE Int. Conf. Mechatronics Automation*, Beijing, China, 2015, pp. 2426-2430.
[5] R. M. Haralick, S. R. Sternberg and X. Zhuang, "Image analysis using mathematical morphology," in *IEEE Trans. Pattern Anal. Mach. Intell.*, vol. PAMI-9, no. 4, pp. 532-550, July 1987.
[6] K. He, X. Zhang, S. Ren and J. Sun, "Deep residual learning for image recognition," in *Proc. IEEE Conf. Comput. Vis. Pattern Recognit.*, Las Vegas, NV, USA, 2016, pp. 770-778.
[7] Y. LeCun, B.E. Boser, J.S. Denker, D. Henderson, R.E. Howard, W.E. Hubbard and L.D. Jackel, "Handwritten digit recognition with a back-propagation network", in *Proc. Advances Neural Information Processing Systems*, Denvor, CO, USA, 1990, pp.396-404.
[8] Y. Lecun, L. Bottou, Y. Bengio and P. Haffner, "Gradient-based learning applied to document recognition," in *Proc. IEEE*, vol. 86, no. 11, pp. 2278-2324, Nov. 1998.
[9] M. Li, T. Zhang, Y. Chen and A.J. Smola, "Efficient mini-batch training for stochastic optimization," in *Proc. 20th ACM SIGKDD Int. Conf. Knowledge Discovery and Data Mining (KDD '14)*, New York, NY, USA, pp. 661-670.
[10] J. Masci, J. Angulo and J. Schmidhuber, "A learning framework for morphological operators using counter-harmonic mean," in *Proc. 11th Int. Symp. Mathematical Morphology Its Appl. Signal Image Process.*, Springer, Berlin, Heidelberg, 2013, pp. 329-340.
[11] G. X. Ritter and P. Sussner, "An introduction to morphological neural networks," in *Proc. 13th Int. Conf. Pattern Recognit.*, Vienna, Austria, 1996, pp. 709-717 vol.4.
[12] G.X. Ritter and J.N. Wilson, "Handbook of Computer Vision Algorithms in Image Algebra," CRC press, 1996.
[13] F. Y. Shih, "Image processing and mathematical morphology: fundamentals and applications," CRC press, 2009.
[14] H. Shih and E. Liu, "Automatic reference color selection for adaptive mathematical morphology and application in image segmentation," in *IEEE Trans. Image Process.*, vol. 25, no. 10, pp. 4665-4676, Oct. 2016.
[15] F.Y. Shih and O. R. Mitchell, "Threshold decomposition of gray-scale morphology into binary morphology," in *IEEE Trans. Pattern Anal. Mach. Intell.*, vol. 11, no. 1, pp. 31-42, Jan. 1989.
[16] J. Stallkamp, M. Schlipsing, J. Salmen and C. Igel, "The German Traffic Sign Recognition Benchmark: A multi-class classification competition," in *Proc. Int. Joint Conf. Neural Netw.*, San Jose, CA, 2011, pp. 1453-1460.
[17] F.Y. Shih, Y. Shen and X. Zhong, "Development of deep learning framework for mathematical morphology," in *Int. J. Pattern Recognit. Artificial Intell.*, vol. 33, no. 6, p. 1954024, June 2019.
[18] K. Simonyan and A. Zisserman, "Very deep convolutional networks for large-scale image recognition," in *Proc. Int. Conf. Learning Representation*, San diego, CA, USA, 2015.
[19] F. Zana and J.C. Klein, "Segmentation of vessel-like patterns using mathematical morphology and curve evaluation," in *IEEE Trans. Image Process.*, vol. 10, no. 7, pp. 1010-1019, July 2001.
[20] W. Zhang, D. Shi and X. Yang, "An improved edge detection algorithm based on mathematical morphology and directional wavelet transform," in *Proc. Int. Congr. Image Signal Process.*, Shenyang, China, 2015, pp. 335-339.